\documentclass{article}

\usepackage{amsmath}
\usepackage{amssymb}
\usepackage{graphicx}
\usepackage[a4paper, left=1in, right=1in, top=1in, bottom=1in]{geometry}
\usepackage{hyperref}
\usepackage{pdfpages}
\usepackage{siunitx}
\usepackage{tabularx,booktabs}
\usepackage{algorithm}
\usepackage{algpseudocode} 

\title{\LARGE \bf
{Design, Calibration, and Control of Compliant Force-sensing Gripping Pads for Humanoid Robots}
}
\date{}

\author{Yuanfeng Han\thanks{Yuanfeng Han is with the Department of Mechanical Engineering, Johns Hopkins University, Baltimore, MD. {\tt\small yhan33@jhu.edu}}, 
Boren Jiang\thanks{Boren Jiang and Gregory S. Chirikjian are with the Department of Mechanical Engineering, National University of Singapore, Singapore. {\tt\small mpegre@nus.edu.sg}}, 
Gregory S. Chirikjian\footnotemark[2]}

\begin{document}

\maketitle
\thispagestyle{empty}
\pagestyle{empty}

\section*{Abstract}
This paper introduces a pair of low-cost, light-weight and compliant force-sensing gripping pads used for manipulating box-like objects with smaller-sized humanoid robots. These pads measure normal gripping forces and center of pressure (CoP). A calibration method is developed to improve the CoP measurement accuracy. A hybrid force-alignment-position control framework is proposed to regulate the gripping forces and to ensure the surface alignment between the grippers and the object. Limit surface theory is incorporated as a contact friction modeling approach to determine the magnitude of gripping forces for slippage avoidance. The integrated hardware and software system is demonstrated with a NAO humanoid robot. Experiments show the effectiveness of the overall approach.

\section{Introduction}
\begin{figure}[t!]
\centering
\includegraphics[width=0.8\linewidth]{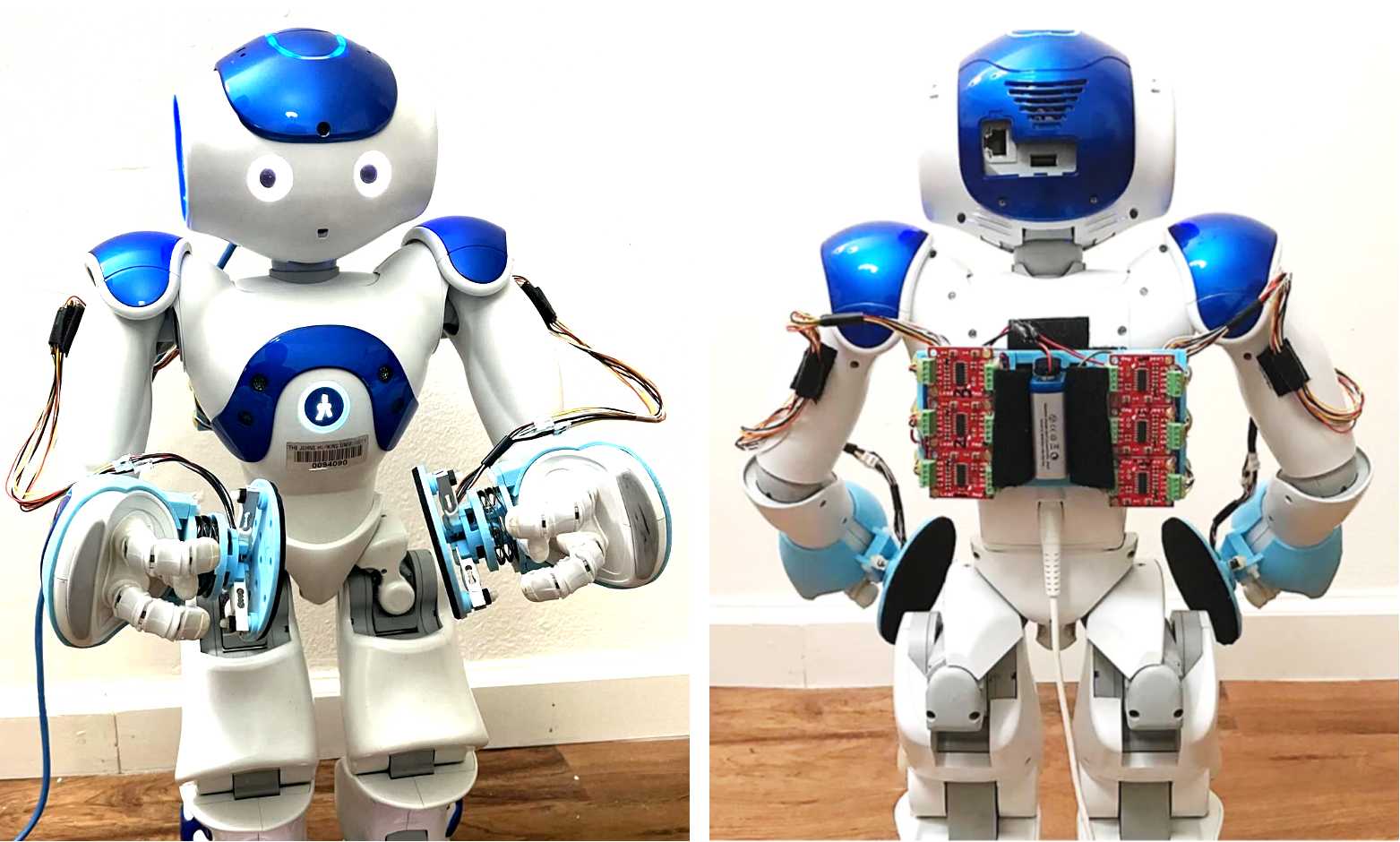}
\caption{Humanoid robot retrofitted with force-sensing gripping pads and associated electronics.}
\label{zero}
\end{figure}

Smaller-sized humanoid robots offer enhanced safety when working close to humans, making them a more affordable and viable option for a wide range of applications, such as future in-home use. Although most smaller-sized and low-cost humanoids are equipped with mature locomotion modules such as walking and step planning \cite{saeedvand2019comprehensive, gouaillier2010omni}, dual-arm manipulation for common rigid box-like objects remains a challenge. This is due to several reasons. First, commercial force/torque (F/T) sensors are often too bulky and expensive for smaller-sized humanoid robots \cite{kim2019multi,cao2021six,ubeda2018design} and many existing customized miniature F/T sensors require complex manufacturing and calibration processes \cite{kim2017surgical,xiong2020six,shams2012compact,kim2020six}. Second, common smaller-sized humanoids possess underpowered motors, leading to imprecise kinematic control of the robot's end-effectors \cite{mattioli2016interaction}. Therefore, the object can easily slip out of the grip during the manipulation process. Third, many current slipping avoidance approaches require the robot to first detect the slippage and then increase the gripping force to resist the slippage \cite{kruse2014sensor}. Such methods often result in slippage accumulation and demand accurate slippage estimation. To address the aforementioned issues, this paper introduces the design of a pair of cost-effective, light-weight and compliant force-sensing gripping pads that can be used for smaller humanoid robots (Figure~\ref{zero}). Furthermore, a gripping force and alignment control framework associated to the gripping pads are proposed to avoid slippage during object manipulation.

 Dual-arm robots have been developed to assist people with various real-world tasks \cite{stephens2010dynamic, arisumi2007dynamic, ohmura2007humanoid}. One of their critical applications is pick-and-place of cumbersome box-like objects in warehouses and factories. A common way to manipulate such objects using dual-arm systems is by first aligning the robot's grippers to the objects' surfaces, applying squeezing forces, and then lifting the object. Grippers designed for handling box-like objects often possess a large flat frictional pad to produce sufficient frictional forces and torques. The pad is often attached to a commercial six-axis F/T sensor for sensory-based control. Gripping pads integrating frictional pads and F/T sensors have been widely applied to dual-arm lifting \cite{han2020can, BostonDynamics}, impedance control \cite{yan2018dual} and teleportation \cite{kruse2014sensor}, etc. Other gripping pad designs also include suction cups \cite{bamotra2019layer, hayakawa2022dual} and gecko adhesive material \cite{suresh2015surface, han2020hybrid, han2022bimanual, roberge2018improving}. Those gripping mechanisms have shown effectiveness in handling flat surfaces and also demonstrated good adaptability to rough or curved surfaces. However, the gecko adhesive material often requires a complex fabrication process \cite{jin2012design} and the gripping surface with this material needs to be aligned properly to produce sufficient adhesion along the direction of gravity \cite{han2022bimanual, roberge2018improving}. Moreover, the suction mechanism requires an air pump to be carried by the robot. Since our goal is to design simple and low-cost grippers for smaller-sized humanoids to manipulate box-like object, we adopt the simple flat frictional pad design with force-sensing capability. 

There are two primary methods for object manipulation utilizing motion and force control: hybrid position-force control \cite{raibert1981hybrid} and impedance control \cite{hogan1985impedance}. The former belongs to the direct force control method. This method examines the measured position and force of the robot's end-effector and uses decoupled position and force feedback control loops to enable the robot to reach its force and position references. Impedance control belongs to the indirect control category, which does not directly steer the robot's position and forces to specific desired values. Instead, it controls the robot's motion to generate a desired dynamic performance for robot-object interaction. For both of the abovementioned methods, the feedback part of the control loop depends on the force and torque measurements from the robot's F/T sensors. 

Another direction of object manipulation using force control lies in the field of parallel grippers \cite{costanzo2019two,chavan2020planar}. These types of grippers have two gripping surfaces parallel to each other, which only allow 1-D motion control to regulate the gripping force. In addition, the surfaces of many parallel grippers are designed with curvature and manufactured using soft materials. When in contact with objects, those surfaces deform and the contact area becomes a flat circular region with axisymmetric pressure distribution. Under this assumption, the required gripping force for slippage avoidance can be solved using the Limit Surface theory \cite{ciocarlie2007soft} (see Section \ref{limit surface theory}). The limit surface sums up the friction across the contact area, providing a more comprehensive contact picture than the Coulomb friction model widely used in robotic manipulation. Nevertheless, such parallel grippers can hardly be used to manipulate large and heavy objects due to their limited contact area.

Although smaller humanoid robots are capable of walking and safely interacting with people, they can hardly be used for object manipulation. One big reason is that they often do not have installed F/T sensors, limiting their knowledge of gripping forces. Commercial F/T sensors widely adopted by industrial dual-arm robots are too expensive and overly heavy, which burdens the robot's actuators. Therefore, it is important to expand the usefulness of the smaller-sized humanoid robots in object manipulation using dual arms by designing light-weight and low-cost force-sensing grippers and developing the associated control methods for the grippers.

The remainder of this paper is organized as follows. Section~\ref{Section_Design} introduces the design of a pair of low-cost and light-weight force-sensing gripping pads. The gripping pads utilize small 1-D load-cells to measure normal force and CoP, and the sensing principle is similar to the force-sensing shoes developed in our previous work \cite{han2021look}. The compliant structure of the gripper facilitates surface attachment and improves the stability of the dual-arm gripping system. Section~\ref{Gripper_Calibration} presents a manual calibration method to improve the gripping pads' CoP measurement accuracy using nonlinear optimization. Section~\ref{limit surface theory} introduces a contact mechanics approach based on the limit surface theory for modeling frictional force and torque between the gripping pads and the objects. Section~\ref{control_design} introduces a kinematic-based hybrid position-alignment-force control method for the gripping pads to manipulate objects in task space. This control method utilizes three decoupled control loops to steer the force, orientation, and position to their corresponding references. The magnitude of gripping forces are determined by the Limit Surface theory to prevent slippage. The grippers and their control methods are tested on a NAO robot and Section~\ref{exp} shows the experiments and results. 

\section{Gripping Pad Design} \label{Section_Design}
\subsection{Design Concept}\label{Section_Design_Concept}
Gripping quality can be evaluated by two measurements: (1) the magnitude of the gripping forces (2) the alignment between the surfaces of the gripping pads and the surfaces of the object. Our force-sensing grippers are designed to directly measure the gripping forces normal to the object surfaces and infer the surface alignment by measuring contact CoP.

\subsection{Mechanical Design}\label{Mechanical Design}
\begin{figure}[t!]
\centering
\includegraphics[width=0.8\linewidth]{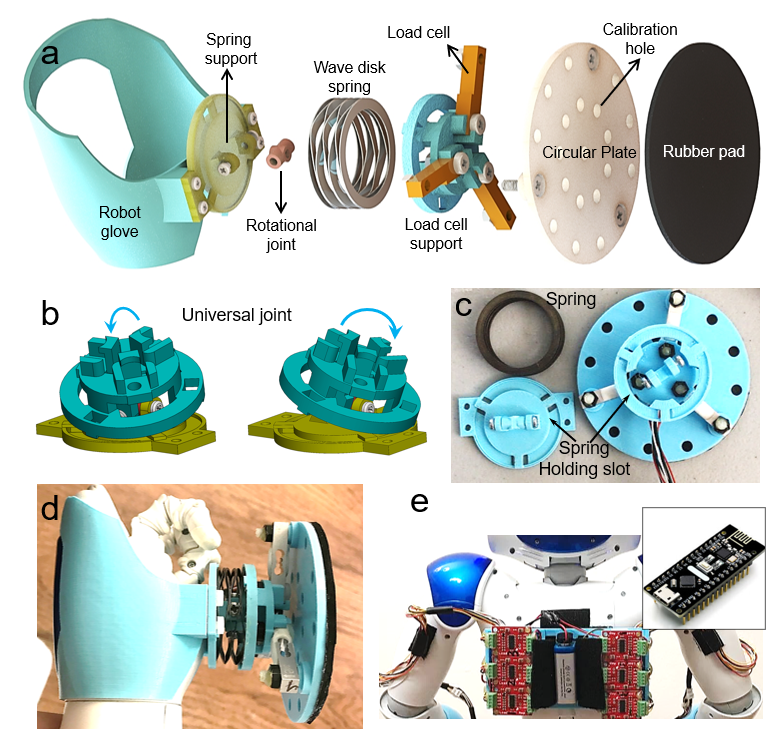}
\caption{(a) Mechanical design of the gripping pad. (b) The universal joint formed using the 3-D printed parts. (c) Disassembled compliant mechanism of the gripping pad. 
(d) A snapshot of a NAO robot wearing a force-sensing gripping pad. (e) Electronic components of the force-sensing gripping pads.}
\label{Design}
\end{figure}

Our force-sensing gripping pads employ small 1-D load cells to measure normal force and CoP. Compared with widely-used light-weight force-sensing resistors and capacitive sensors, load cells with a linear voltage-to-force response are more accurate. Each gripper consists of three load cells fixed on a 3-D printed support (Figure~\ref{Design}a). This support, together with a 3-D printed base plate, are connected to a rotational joint with two orthogonal holes (Figure~\ref{Design}a). These three parts form a universal joint with a maximum rotation angle of 25$^{\circ}$ (Figure~\ref{Design}b). A stacked wave disk spring is utilized to improve gripper-object alignment (Figure~\ref{Design}a). The spring is fixed into the grooves between the base and the support (Figure~\ref{Design}a and c). The wave disk is chosen due to its limited torsional movement compared with traditional coil-type springs, protecting the universal joint structure from the impact of the gripping. A 3-D printed circular plate with a frictional rubber pad is mounted on the tips of the load cells (Figure~\ref{Design}a). The plate is manufactured with circular-patterned holes used for calibration (Figure~\ref{Design}a). All the components of the gripper are assembled and fixed onto a 3-D printed glove worn by the NAO (Figure~\ref{Design}d). Different from many existing grippers with rigid structural design, our compliant gripping pads can be theoretically proved to passively stabilize the gripping system \cite{calanca2016role}. Each gripper weights 35g and its components (including electronics) cost less than 45 US dollars, which means that it is affordable enough to be widely used in many smaller-sized and low-cost dual-arm platforms. The mechanical parts of the grippers are listed in Table~\ref{GripperParts}.

\begin{table}[ht]
\caption{List of Parts and Vendors}
\centering
\begin{tabularx}{1.0\textwidth} { 
   |>{\centering\arraybackslash}X | 
    >{\centering\arraybackslash}X |
    >{\centering\arraybackslash}X |
    >{\centering\arraybackslash}X| }
 \hline
Part & Specification/Material & Vendor/Manufacturer   \\
\hline
Plate                    &  PLA             & Amazon      \\
Plastic screws/nuts       &  M4              & Mcmaster Carr, NJ\\ 
Stacked wave disk spring  &  7756N692        & Mcmaster Carr, NJ\\ 
Load cell                &  BF-02088 B      & Bingf STCL  \\
ADC                       & Sparkfun HX711   & Sparkfun    \\
MCU+wireless sender       & RF-Nano          & Amazon      \\
Wireless receiver module         & NRF24L01         & Amazon      \\
Receiver MCU                       & Arduino Mega     & Amazon      \\
\hline      
\end{tabularx}
\label{GripperParts}
\end{table}
\subsection{Electronic Design}
The electronic components of the gripping pads are chosen off-the-shelf (Table~\ref{GripperParts}). Three modules are developed to acquire force data, including a transmission module, a receiving module, and a processing module. In the transmission module carried on the robot's back (Figure~\ref{Design}e), the load cells' voltage outputs pass their corresponding analog-to-digital converter (ADC) and are transmitted using a micro-controller-unit (MCU) with an embedded wireless module (Figure~\ref{Design}e). Then a sister wireless receiver connecting to an MCU receives the signal and transfers it to a PC via serial communication. Finally, the processing module, established in the robotics operating system (ROS), denoises the data using median filters and outputs data at 85 Hz.

\subsection{Force-sensing Principle}
\begin{figure}[t!]
\centering
\includegraphics[width=0.4\linewidth]{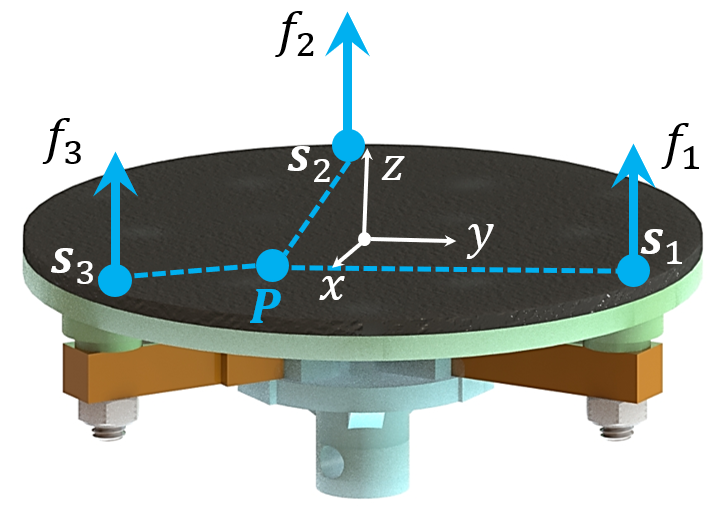}
\caption{Schematics of the measured normal force and CoP of the gripping pad.}
\label{Principle}
\end{figure}

As introduced in Section~\ref{Section_Design_Concept}, each gripping pad measures normal force (NF) and contact CoP. The latter represents the averaged pressure center and is a direct indication of how well the flat frictional pad is aligned with the surface of a box (see Section \ref{limit surface theory}). Denoting the measured force {vector} of each load cell as $\bf f_{i}$ (Figure~\ref{Principle}), the normal gripping force vector ${\bf F}$ is the sum of all the measured force vectors: 
\begin{align}
{\bf F} = \sum_{i = 1}^{3}{\bf f_{i}}.\label{normal_force}
\end{align}
The CoP describes a point where the normal force has zero resultant moments \cite{kajita2014introduction}. Here we first calculate the total moment of all the normal forces at a random point ${\bf p}$ (Figure~\ref{Principle}):
\begin{align}
{\boldsymbol\tau}({\bf p}) =  \sum_{i = 1}^{3}({\bf s_{i}} - {\bf p})\times {\bf f_{i}},
\end{align}
where ${\bf s_{i}}$ is the coordinate of the sensor $i$ in the pad's body frame (Figure~\ref{Principle}).The value of {\bf p} such that ${\boldsymbol\tau({\bf p})} = {\bf 0}$ is denoted as ${\bf p_{c}}$, which is the CoP. Since forces are normal to the plane, the first and second elements of the total moment are set to zero to solve for ${\bf p_{c}}$:
\begin{align}
{\bf p_{c}} = \sum_{i=1}^{3}||{\bf f_{i}}||{\bf s_{i}}/ \sum_{i=1}^{3}||{\bf f_{i}}||.
\end{align}
If the CoP is at the pad's geometric center, the contact pressure evenly distributed across the surface of the gripper (considering the surfaces of the gripping pad and the object are both rigid). If the CoP is very close to the edge of the pad, either the pad is close to pivot around the edge or it has already lost full contact.

\begin{figure}[t!]
\centering
\includegraphics[width=0.6\linewidth]{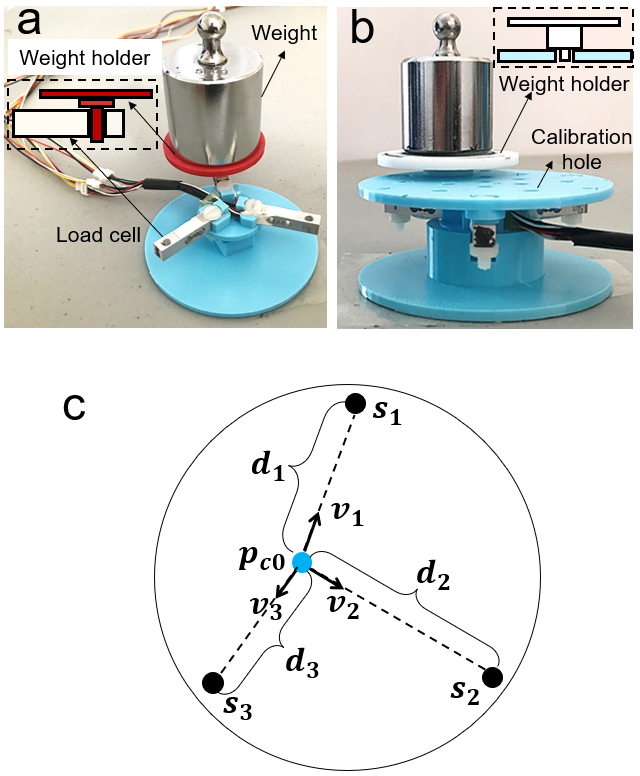}
\caption{(a) Load cell calibration. (b) Force-sensing gripping pad CoP calibration setup. (c) CoP calibration parameters.}
\label{Gripper_Cal}
\end{figure}
\section{Gripping Pad Calibration and Evaluation} \label{Gripper_Calibration}
\subsection{Load Cell Calibration}
For the grippers to function, each load cell is calibrated to convert its voltage output to the physical force. This calibration is implemented by applying a known weight to the load cell's measuring point and recording the corresponding voltage change (Figure~\ref{Gripper_Cal}a). The averaged unloaded and loaded voltages of the load cell are denoted as $S_{0}$ and $S_{G}$. Due to the load cell's linear voltage-to-force response, the voltage-to-force ratio $b$ can be calculated by dividing the voltage change $(S_{F}-S_{0})$ by the weight of the calibration mass, $G$:
\begin{align}
b = \frac{S_{G}-S_{0}}{G}.    
\end{align}
The mapping from the sensor's voltage reading $S_{F}$ to its corresponding force output $f$ (\ref{normal_force}) is given by:
\begin{align}
f = \frac{(S_{F}-S_{0})}{b} = \frac{S_{F}}{b} - \frac{S_{0}}{b} = cS_{F}+d, \label{affine}
\end{align}
where $c$ and $d$ are the modified scaling factor and offset.

\begin{figure}[t!]
\centering
\includegraphics[width=0.85\linewidth]{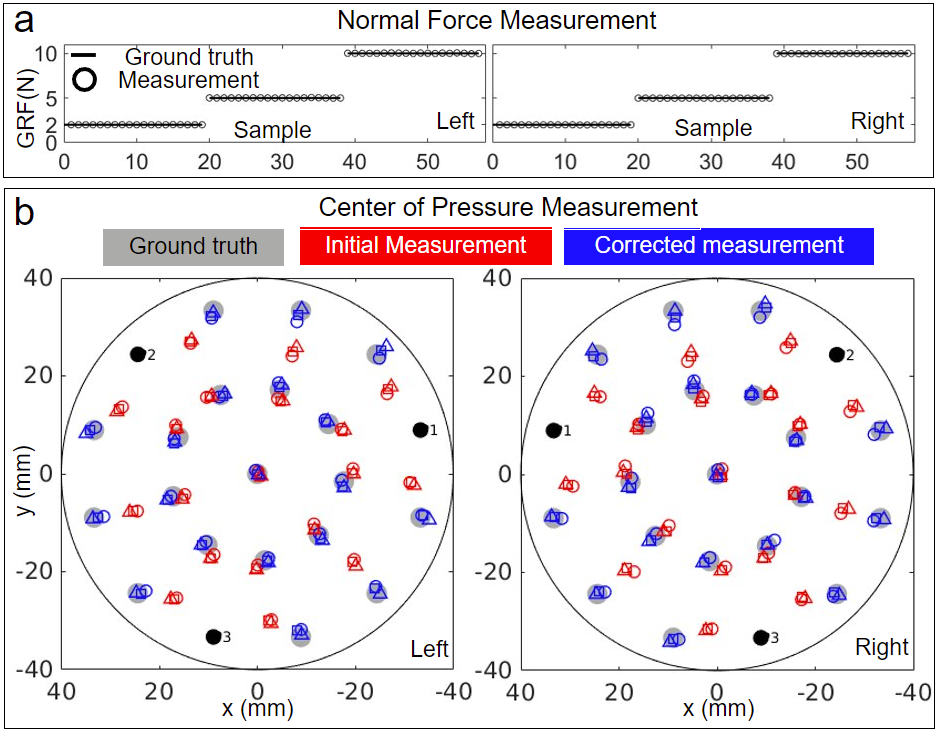}
\caption{(a) Normal force measurement precision of the force-sensing gripping pads, respectively. Left and right plots show the result for the left and right gripping pads. Black lines show the ground truth for measurement using 0.2, 0.5, 1.0 kg weights. Circles represent measurement using each weight at different calibration positions. (b) CoP measurements precision of the force-sensing grippers. Left and right plots show the left and right gripping pads. Gray, red and blue markers show CoP ground truth and CoP measurements before and after calibration. Circular, square and triangle markers show measurements using 0.2, 0.5, 1.0 kg weights. Black circles show the sensor locations.}
\label{Gripper_Calibration_Result}
\end{figure}

\subsection{Measurement Evaluation}
We evaluated the measurement precision of the calibrated load cells by 
placing a 0.2 kg, 0.5 kg, 1.0 kg mass on each of the 19 holes of the pad's circular plate using a flat holder (Figure~\ref{Gripper_Cal}b). The normal forces and CoPs (Figure~\ref{Gripper_Calibration_Result}) are measured and compared with the known weight of the loads and the $x-y$ coordinates of the holes loading the weights. We concluded that the measured normal forces (Figure~\ref{Gripper_Calibration_Result}a, circles) almost perfectly lie on their correspondence (Figure~\ref{Gripper_Calibration_Result}a, black lines), with the mean absolute error (MAE) only around 0.02N (Table~\ref{GripCalTable}). However, the CoPs (Figure~\ref{Gripper_Calibration_Result}b, red markers) deviate from their correspondence (Figure~\ref{Gripper_Calibration_Result}b, gray circles). The error is likely caused by the mechanical misalignment and internal deformation between the load cells and the plate. Therefore, the measured CoPs need further calibration.

\subsection{CoP Calibration}
As shown in Figure~\ref{Gripper_Calibration_Result}b, the measured CoPs at each location corresponding to different weights (red circles, squares and triangles corresponding to 0.2, 0.5 and 1.0 kg weights) are tightly clustered. This indicates that the magnitude of normal force does not significantly impact the CoP measurement. However, the measured CoPs far from the pad's geometric center (red markers at the outer circle) tend to deviate from their correspondence (gray dots) by moving towards their nearest sensors (black dots). This observation motivates us to use the relative location between the measured CoPs and the sensors as the parameters for CoP correction. Specifically, two correction terms $\Delta x$ and $\Delta y$ are applied to each measured CoP, ${\bf p_{c0}}$, and the corrected CoP, ${\bf p_{c}}$, is given by:
\begin{align}
    \begin{bmatrix}
    p_{c\_x}\\
    p_{c\_y}
    \end{bmatrix} = 
    \begin{bmatrix}
    p_{c0\_x} + \Delta x\\
    p_{c0\_y} + \Delta y
    \end{bmatrix}. 
\end{align}
We denote a unit vector pointing from ${\bf p_{c0}}$ to the $i$th sensor, ${\bf s_{i}}$, as ${\bf v_{i}}$, and the distance between ${\bf p_{c0}}$ and ${\bf s_{i}}$ as $d_{i}$ (Figure~\ref{Gripper_Cal}, c). The correction terms are defined by the vector sum of all the unit vectors, which are weighted by a third order polynomial parameterized by $d_{i}$: 
\begin{align}
    \begin{matrix}
        \Delta x =  \sum_{i=1}^{3}\sum_{j=1}^{3}a_{ij}{d_{i}}^{j-1}v_{i\_x}\\
        \Delta y =  \sum_{i=1}^{3}\sum_{j=1}^{3}b_{ij}{d_{i}}^{j-1}v_{i\_y}\\
    \end{matrix}
\end{align}
in which $i$ and $j$ represents the $i$th load cell and the $j$th order of the polynomial. The parameters $a_{ij}$ and $b_{ij}$ define the polynomials, which can be solved by matching the corrected CoPs, ${\bf p_{c}}$, (Figure~\ref{Gripper_Calibration_Result}b, red markers) to their correspondence ${\bf p_{g}}$ (Figure~\ref{Gripper_Calibration_Result}b, gray circles) using nonlinear least squares:
\begin{align}
     \underset{a_{ij}, b_{ij}}{\text{argmin}} \  J  = \sum_{k=1}^{n}||{\bf p_{c}}[k] - {\bf p_{g}}[k]||^2,
\end{align} where $k$ represents the sampled measurements. 

As shown in Figure~\ref{Gripper_Calibration_Result}b, the corrected CoPs after calibration (blue markers) align much closer to their corresponding ground truths (gray markers) compared to the initial measurements (red markers). The MAE of the corrected CoP is approximate 1 mm, which is significantly improved compared to that of the initial measurements (Table~\ref{GripCalTable}). The result shows that our calibration method effectively improves the CoP measurement accuracy.

\begin{table}[ht]
\caption{Mean Absolute Error}
\centering
\begin{tabularx}{0.7\textwidth} { 
    |>{\centering\arraybackslash}X | >{\centering\arraybackslash}X 
    >{\centering\arraybackslash}X 
    >{\centering\arraybackslash}X| }
 \hline
& CoP Measured (mm) & CoP Corrected (mm) & Normal Force Measured (N)  \\
 \hline
 Left Pad  & 5.00 $\pm$ 2.98 & 1.00 $\pm$ 0.54 & 0.017 $\pm$ 0.014 \\
\hline
 Right Pad  & 4.75 $\pm$ 2.82 & 0.99 $\pm$ 0.43 & 0.013 $\pm$ 0.018    \\
\hline
\end{tabularx}
\label{GripCalTable}
\end{table}

\section{Contact Analysis}\label{limit surface theory}
Our gripping control law is based on the contact friction modeling introduced in this section. Although many existing studies in dual-arm manipulation model contact friction using friction cone theory, this theory is insufficient to quantify the frictional torque for grippers with large contact areas. 
Our interest in this paper focuses on flat surface interaction between the rigid surface of our frictional pads and a rigid box-like object. Here we do not consider cases where interaction causes large surface deformations. To better model the contact friction forces and torques, we adopt Limit Surface Theory \cite{goyal1991planar, howe1996practical} to quantify the friction in flat surface interactions. 

\subsection{Contact Mechanics} \label{contact mechanics}
\begin{figure}[t!]
\begin{center}
\includegraphics[width=0.75\linewidth]{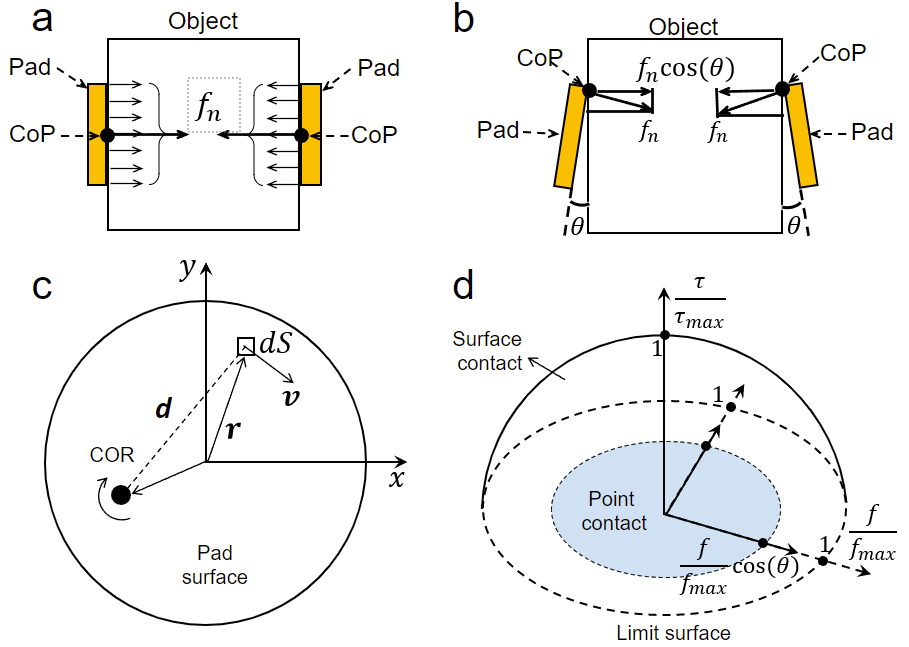}
\caption{(a) and (b) show the pressure distribution and CoP of a well-aligned (without external moments acting on the pads) and a miss-aligned contact. (c) A contact in the sliding surface of the force-sensing gripping pad. The contact is rotating clockwise about the instantaneous CoR, with resulting direction of velocity ${\bf \hat{v}} = {\bf v}/||\bf v||$  at the infinitesimal contact area $dS$. (d) Two limit surfaces: the outer transparent hemisphere represents the limit surface of case (a) and the 2D blue circle represents case (b).}
\label{ContactMechanics}
\end{center}
\end{figure}

 The contact mechanics of two typical contact configurations in object gripping are analyzed. Case 1: the flat rigid pad is well attached to the rigid surface of a box (Figure~\ref{ContactMechanics}a). Case 2: the pad is not in full contact with the surface of the box (Figure~\ref{ContactMechanics}b).
To analyze the interaction motion between the pad and the box for the above two cases, we first model the instantaneous motion of the pad in a plane using a pure rotation about a certain location, known as the center of rotation (CoR) \cite{howe1996practical}. By assuming a given CoR and then adding the contribution of the friction force at each point throughout the contact area, one can figure out the relationship between sliding motion and the total friction force and moment for the sliding body. The vector from a given CoR $(x_c,y_c)$ to an infinitesimal area $dS(x,y)$ in the contact zone is ${\bf d} = [x-x_c,y-y_c]^T$ (Figure~\ref{ContactMechanics}c). The unit velocity of each infinitesimal area is denoted as ${\bf v}$ with resulting direction of velocity ${\bf \hat{v}} = {\bf v}/||\bf v||$ (Figure~\ref{ContactMechanics}c). Since the velocity vector is perpendicular to $\bf d$, we have

\begin{align}
    {\bf \hat{v}} = \frac{[y-y_c,x_c-x]^T}{\sqrt{(y_c-y)^2+(x-x_c)^2}}
\end{align}
The magnitude of the friction
force vector of an infinitesimal area is $d{\bf f_t} = \mu p(x,y) dA $, where $\mu$ is assumed to be a constant coefficient of friction, and $p(x,y)$ is the local value
of the pressure distribution. Because the friction force is in the opposite direction to the relative motion of the point, the local friction force vector is $ d{\bf f_t} = -\mu p(x,y) {\bf \hat{v}} dA $. The total tangential friction is calculated by summing up the contribution of the frictional force at each infinitesimal area $d\mathbb{S}$ across the contact \footnote{We consider objects with rigid surfaces. Therefore, the integration area $\mathbb{S}$ is the overall surface of the gripping pad.}:
\begin{align}
{\bf f_{t}} = -\int_{\mathbb{S}}\mu{p(x,y)}\hat{{\bf v}}(x,y)d{\mathbb{S}}.
\label{friction}
\end{align}
The friction moment of a point in the contact area is given by the cross-product of the vector $\bf r$ and the local friction force as $ d{\bf \tau_t} = -({\bf r}\times {\bf \hat{v}}) \mu p(x,y) dA $, where $\bf r$ is the vector from the origin to an infinitesimal area (Figure~\ref{ContactMechanics}c). The total frictional torque can be obtained by summing up the contribution of the frictional moment of each point:
\begin{align}
\boldsymbol\tau = -\int_{\mathbb{S}}({\bf r}\times\hat{{\bf v}}(x,y))\mu{p}(x,y)d\mathbb{S}
\label{moment}
\end{align}
The relationship between sliding motion and force-moment can be obtained by solving these equations for a number of CoR positions. The limit surface of this case (Figure~\ref{ContactMechanics}a) is shown in Figure~\ref{ContactMechanics}d with the
friction force forming the horizontal plane and the friction torque forming the vertical axis. The maximum tangential force ${\bf f_{\text{max}}}$ takes place in pure translational sliding, where ${\bf \hat{v}(x,y)}$ becomes a constant and
\begin{align}
{\bf f_{\text{max}}} = -\mu{\bf \hat{v}}\int_{\mathbb{S}}{p(x,y)}d{\mathbb{S}} = -\mu{f_n}{\bf \hat{v}},
\end{align}
where ${\bf f_{n}}$ is the total normal force of the pad. Here the limit surface determines whether a sliding motion will take place when a single pair of tangential force and moment is applied to the contact surface of an object \cite{howe1996practical}. If the point of frictional force and moment combination is inside the surface, no sliding motion will occur; if the point is on the surface, a steady sliding motion will happen; and if the point is outside the surface, a sliding acceleration of the object will occur, due to the fact that the applied force and moment beyond the friction forces of steady sliding. However, there is no analytical solution for the limit surface except for some special pressure distributions.

Since this paper focuses on gripping control between flat, rigid, non-deformable surfaces, instead of adopting the Hertzian pressure distribution, which is widely used by soft deformable fingers \cite{xydas1999modeling} or parallel grippers \cite{costanzo2019two}, we apply a traditional contact mechanics modeling method called dimensionality reduction (MDR) \cite{popov2019handbook}, which is a general approach designed for axially symmetric contact. This method describes three-dimensional contact by modeling the interaction forces between the surface and a one-dimensional array of independent springs. Therefore, when the gripping pad is well attached to a rigid surface without external moment applying to it (Figure~\ref{ContactMechanics}a), due to the symmetric shape of our circular pad, the rigidity of the pad surface and the MDR elastic formulation, we can assume a uniform pressure distribution between these surfaces and the averaged pressure center (CoP) is located at the geometric center of the pad (Figure~\ref{ContactMechanics}a). 

\begin{figure}[ht]
\centering
 \includegraphics[scale=0.35]{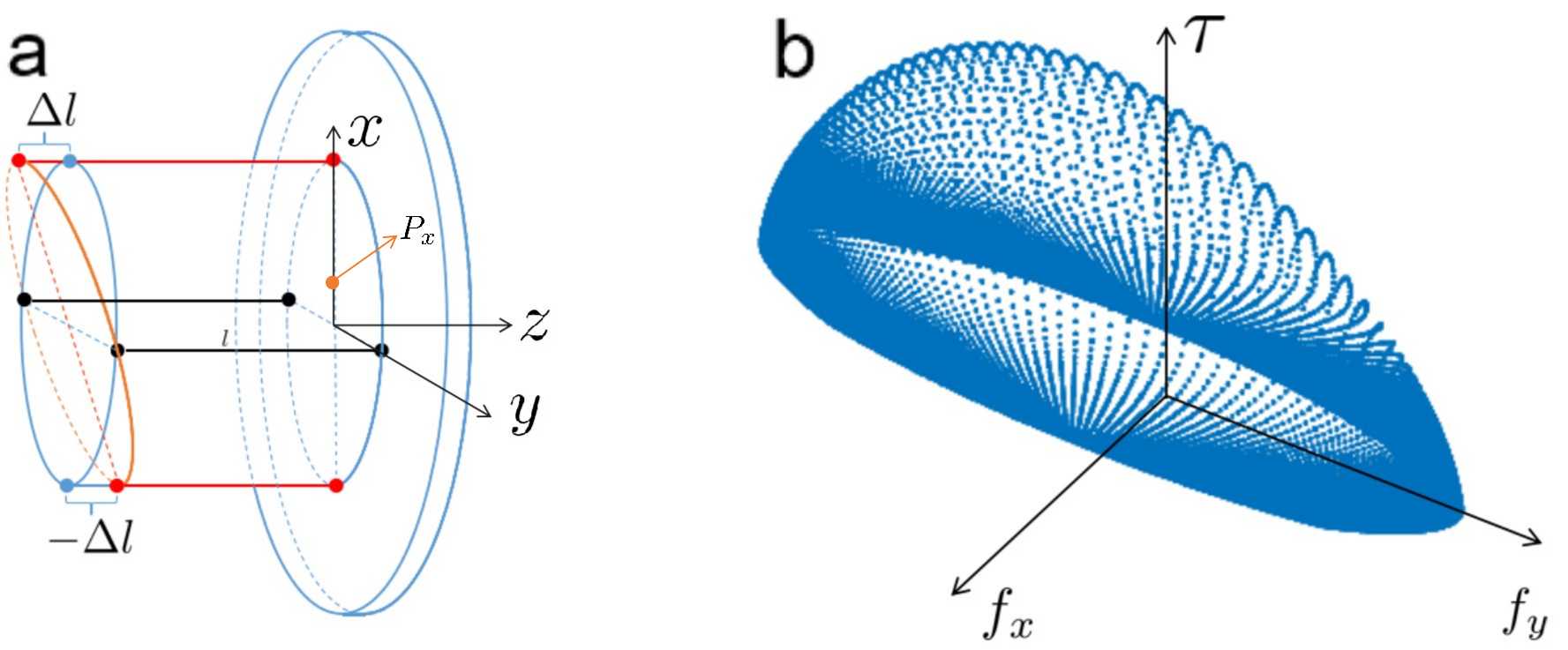}
 \caption{(a) The deformation of the wave disk spring when external torque is applied to the gripping pad. (b) The profile of a simulated limit surface for the case in (a).}
 \label{pre}
\end{figure}

When an external moment is applied to the pad through the wave disk spring connected to it (Figure~\ref{Design}a), according to Hooke's law, the pressure is proportional to the change in length $\Delta l$ of the wave disk spring. For example, when the universal joint (Figure~\ref{Design}b) rotates along the $y-$axis direction in Figure~\ref{pre}a, the lengths of two red lines along the $x$-axis become $l+\Delta l$ and $l-\Delta l$, and the length of two black lines along the $y$-axis remain $l$. Therefore, the resulting pressure distribution is given by:
\begin{align}
p(x) = ax+p_0,
\label{pd}
\end{align}
where the averaged pressure $p_{0} = f_{n}/\pi R^{2}$, and the CoP, $\bf{P}_x$, is on the $x$-axis (Figure~\ref{pre}a). Accordingly, a new simulated limit surface can be visualized in Figure~\ref{pre}b. In general, as the CoP moves around the gripping pad with the same distance to the pad's geometric center, the limit surface will rotate around the $\tau$-axis (Figure~\ref{pre}b) with the same profile and the maximum moment does not change. This property will be applied to formulate our control law in Section \ref{force regulation}.

For Case 2, if the CoP is on the edge of the pad, this indicates that the contact between the pad and the object is about to break or the pad has already tilted away from the surface with only its edge in contact (Figure~\ref{ContactMechanics}b). For the latter case, the contact can be considered as a single frictional point, which does not produce frictional torque. The limit surface of this case is approximated as a 2D circle (Figure~\ref{ContactMechanics}d, blue circle), of which the maximum tangential force is:
\begin{align}
{\bf f_{\text{max}}} = -\mu{f_n}{\bf \hat{v}}\cos(\theta),
\end{align}
where $\theta$ is the tilting angle between the pad and surface (Figure~\ref{ContactMechanics}b). 

Comparing the above cases, we see that the limit surface with a well-aligned pad can resist substantial gravitational force and torque (Figure \ref{ContactMechanics}d), while a miss-aligned pad can easily lose resistance to gravitational torque leading to rotational slippage (Figure~\ref{ContactMechanics}b, blue circle). In our control method in the next section, we design a strategy to control the pad's CoP close to its geometric center by adjusting the pad's orientation.

\section{Control} \label{control_design}
\subsection{Control System Overview}
\begin{figure}[ht]
\centering
\includegraphics[width=0.75\linewidth]{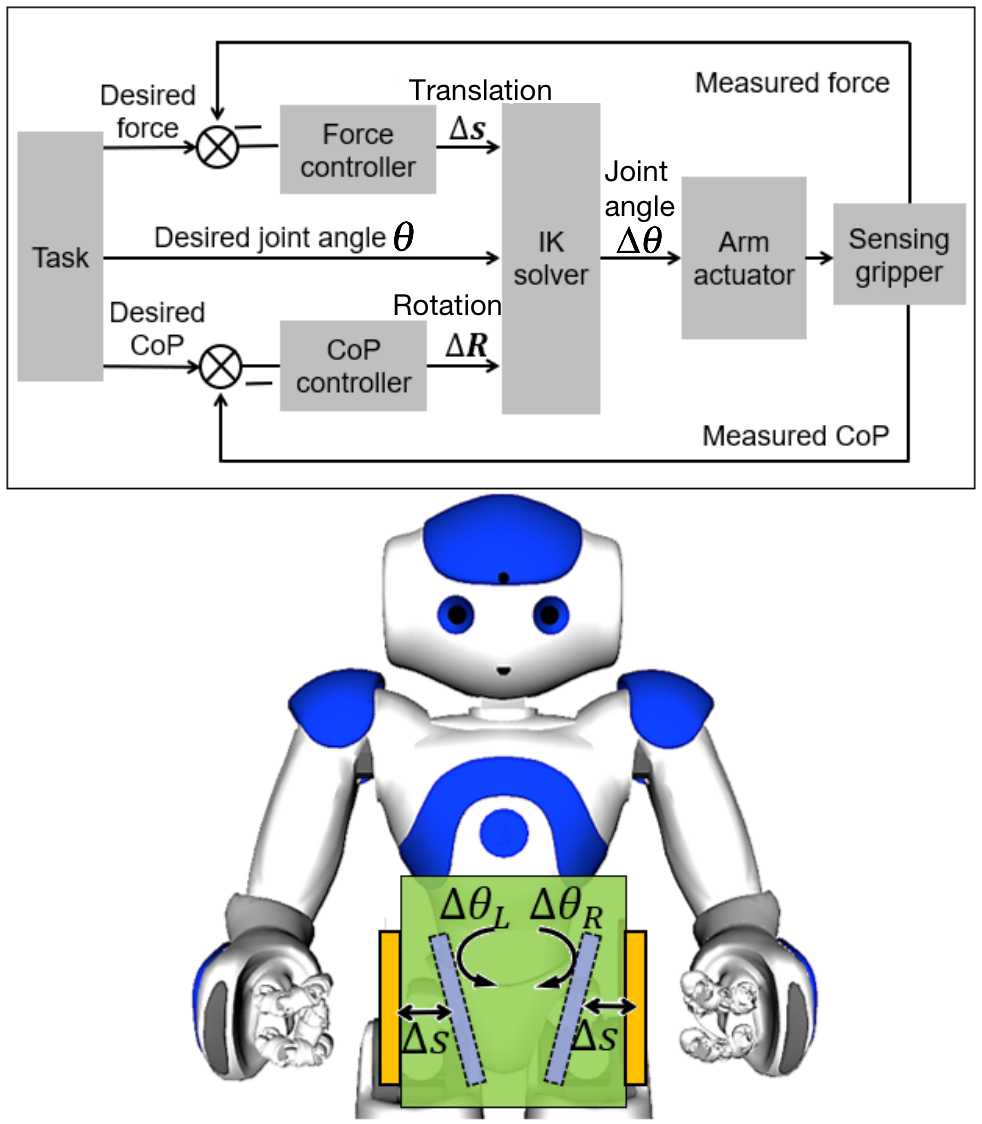}
\caption{Top: the control diagram of our position-alignment-force controller. Bottom: schematics of the kinematic-based control law. $\Delta S$ and $\Delta \theta$ are the control outputs of force and alignment control loops. Orange and gray rectangles represent the actual position and orientation of the gripping pads and their commanded position and orientation from the controllers.}
\label{Controller}
\end{figure}

The control system for dual-arm manipulation utilizes a modified hybrid position-force control method, which comprises three decoupled control loops (Figure~\ref{Controller}, top): 1. A task space controller generates a feasibly dual-arm trajectory (a series of arm joint configurations) for a manipulation task. 2. An alignment controller adjusts the orientation of the gripping pad based on the distance between the measured CoP and the pad's geometric center to facilitate pad-object surface attachment. 3. A force controller adjusts the commanded distance between the pad' geometric centers according to the difference between the measured gripping normal force and its reference. Our control method is established upon two assumptions: 1. The magnitude of the stable initial gripping force to prevent the box from slipping at its current pose is a priori known. 2. The center of mass (CoM) position of the box in its body frame is a priori known. Obtaining the initial stable gripping force and estimating the box's CoM are beyond the discussion of this paper. However, practically, a robot can use an iterative search method to find a proper initial lifting gripping force and then estimate the box's CoM property using equipped force sensors as described in our previous work \cite{han2020can}.

\subsection{Task Space Controller}
The task space controller determines the robot's arm trajectory using the following steps. First, we run the force and CoP controllers (Figure~\ref{Controller}, top) to enable the robot to reach its initial stable gripping force and CoP references. Then, the resulting commanded distance between the pads' geometric centers and the commanded orientations of the pads by the controllers are recorded (Figure~\ref{Controller}, bottom, gray rectangles). Then, the planner generates a dual-arm trajectory, kinematically preserving the recorded distance and orientations of the pads. The trajectory consisting of a series of joint configurations is solved using the direct collocation method given by:
\begin{align}
\underset{{\bf  q},{\bf  u}}{\text{minimize}} \quad & J = \sum_{k=1}^{N}( ||{\bf  q}_{\text{goal}}-{\bf  q}[k]||^{2}_{{\bf  Q_{q}}} + ||{\bf  u}[k+1] - {\bf  u}[k]||^{2}_{{\bf  Q_{u}}}) & \text{(Cost function)} \label{15}\\
\textrm{subject to:} \quad 
&{\bf  q}[1] = {\bf  q}_{\text{i}}, \quad {\bf  q}[N] = {\bf  q}_{\text{g}} & \text{(Initial/goal configuration)} \label{16}\\
&{\bf  q}[k+1] = {\bf  q}[k] + {\bf  u}[k] & \text{(state  transition)}\label{17}\\
&{\bf  T_{L}}[k] - {\bf   T_{R}}[k] = [0,\Delta s,0]^{\mathsf{T}} & (\text{Relative translation})\label{18}\\
&{\bf  R_{L}}[k] = {\bf  R_{L}}[1] & (\text{Left orientation}) \label{orient1}\\
&{\bf  R_{R}}[k] = {\bf  R_{R}}[1] & \text{(Right orientation}) \label{orient2}\\
&{\bf  q}_{\text{min}} < {\bf  q}[k] < {\bf  q}_{\text{max}} & (\text{Joint angle limit})\label{21}\\
&{\bf  d}[k] = {\text{Dist}}{({\bf  q}[k])} < {\bf   d_{\text{min}}} & (\text{Collision avoidance})\label{22}\\
& \text{Kinematics constraints} \label{23},
\end{align}
where $k$ is the node index of the trajectory. In this problem, the optimization variables are joint configuration vectors, $\mathbf{q}$, and control vectors, $\mathbf{u}$, along the trajectory. The control vectors $\bf{u}$ are designed to connect the consecutive joint configurations (\ref{17}). The cost function \ref{15} penalizes the distance between the configuration along the path and the goal configuration to drive the robot arm to its goal configuration and also minimizes the control difference to smooth the trajectory. The constrain sets include the initial and goal configuration constraints (\ref{16}), the distance between the grippers (\ref{18}), the orientation of the grippers (\ref{orient1}-\ref{orient2}) (the index 1 means the initial orientation of the pads) and other kinematics constraints including joint limits and collision avoidance (\ref{21}-\ref{23}). The generated trajectory is a high-level guidance for the motion of the robot arms, which does not incorporate alignment and force adjustment.

\subsection{Alignment Controller} \label{alignment control}
The alignment controller adjusts the orientation of the pad according to the measured CoP location. The goal of the alignment control is to maintain the CoP close to the gripper's geometric center. A measured CoP that deviates from the gripper's geometric center indicates an increased pressure distribution along the direction pointing from the geometric center to the CoP (Figure~\ref{OrientControl}a). As a result, a reversed rotation of the gripper is required to distribute the pressure evenly. Specifically, denoting the measured CoP, ${\bf p_{c}} = [p_{x}, p_{y}]^{T}$, as a vector defined in the pad's local frame (Figure~\ref{OrientControl}b), ${\bf n} = [n_{x}, n_{y}]^{T}$, as a unit vector, which is also the axis of the counter-rotation (Figure~\ref{OrientControl}b) obtained by rotating $\bf p_{c}$ around the z-axis of the pad by 90$^{\circ}$: 

\begin{figure}[t!]
\centering
\includegraphics[width=0.70\linewidth]{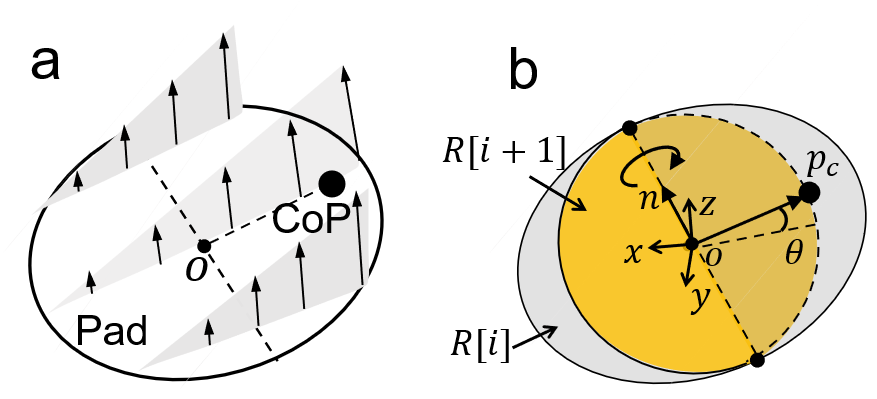}
\caption{(a) Pressure distribution of a rigid frictional pad with CoP not in the pad's geometric center. (b) The generated axis ($\bf{n}$) and angle ($\theta$) for a counter rotation according to the measured CoP ($\bf{p_{c}}$). $R[i+1]$ and $R[i]$ represent the updated orientation from the controller and the previous orientation of the pad.}
\label{OrientControl}
\end{figure}

\begin{align}
\begin{bmatrix}
n_{x} \\
n_{y} \\
1
\end{bmatrix} = 
\begin{bmatrix}
0 & -1 & 0\\
1 & 0 & 0 \\
0&0&1
\end{bmatrix}
\begin{bmatrix}
p_{x} \\
p_{y} \\
1
\end{bmatrix}
\end{align}
An integral control law is adopted to adjust the rotation angle, $\Delta \theta$, based on the distance between the CoP and the pad's geometric center (Figure~\ref{OrientControl}b):
\begin{align}
\Delta\theta = K_{c}g(||{\bf p_{c}}||)\quad (K_{c}>0).
\end{align}
where $K_{c}$ is the gain of the integral controller. $g(.)$ is a dead zone with a threshold of 10 mm, which prevents smaller errors from generating unwanted oscillation.
The relative rotation is given by a rotation matrix defined using Rodrigues' formula:
\begin{align}
    \mathbf{R}({\bf  n},\Delta\theta) = \mathbf{I} + \hat{{\bf  n}}\sin{(\Delta\theta)} + \hat{{\bf  n}}^{2}(1-\cos{(\Delta\theta)}),
\end{align}
with
\begin{align}
\hat{{\bf   n}} =  \begin{bmatrix}
0 & 0 & n_{y} \\
0 & 0 & -n_{x} \\
-n_{y} & n_{x} & 0
\end{bmatrix}.
\end{align}
The updated rotation ${\bf  R}[i+1]$ after each alignment control loop is given by:
\begin{align}
\mathbf{R}[i+1] = \mathbf{R}[i]\mathbf{R}({\bf  n},\Delta\theta)
\end{align}

\subsection{Direct Force Control}
Our gripping controller regulates both force and alignment in decoupled loops. We first introduce the force control loop. Initially, the frictional pad moves towards a commanded position parallel to the object's surface while maintaining a small gap. Then the pads move toward the surfaces along the horizontal direction until the desired force is reached. The left and right arms always move an equal amount of distance $\Delta s$ (Figure~\ref{Controller}, bottom) to maintain the box's horizontal position, and the averaged force of the left and right arms is used as the feedback. As suggested by [12] on robustness integral force control using a robotic arm, we adopt the integral control law. Therefore, the change of position per unit time $\Delta s$ is given by:

\begin{align}
\Delta s = K_{f}{g_{f}}\left(\frac{1}{2}(f_{L} + f_{R}) - f_{d}\right),
\end{align}
where $K_{f}$ is the integral gain, $f_{L}$ and $f_{R}$ are measured forces of the left and right arms, $f_{d}$ is the desired force, $g(.)$ is a dead zone which prevents smaller errors from generating unwanted oscillations. To prevent the integral controller from winding up between the time when the controller starts and the time the pad contacts the surface, a variable desired force is designed:
\begin{align}
     f_{d} =
    \begin{cases}
      f_{d1} & f_{n} \leq \xi\\
      f_{d2} & f_{n} > \xi
    \end{cases} \quad\quad (f_{d1} < f_{d2})   
\end{align}
where $\xi$ is a small force.

\subsection{Gripping Force Regulation} \label{force regulation}
\begin{figure}[t!]
\centering
\includegraphics[width=0.8\linewidth]{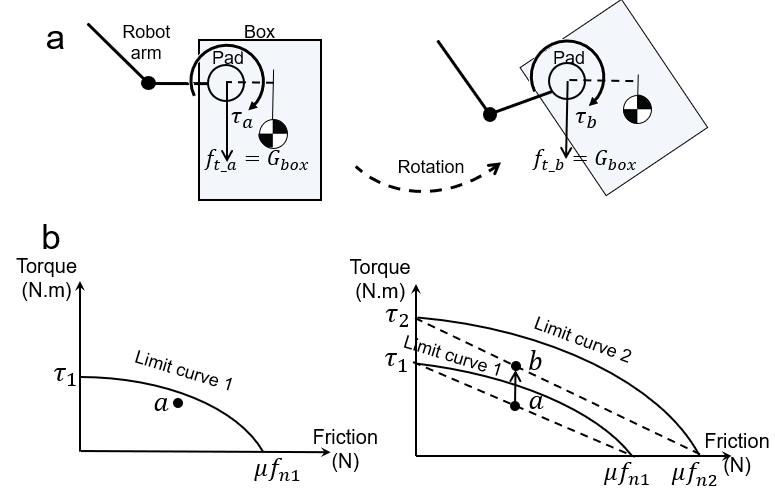}
\caption{Gripping control based on limit surface theory. (a) Left: frictional force and torque analysis to balance a box lifted up vertically. Right: force and torque analysis for a rotated box. (b) Left: a limit curve to avoid slippage for the vertical box. Right: the expanded limit curve to avoid slippage for the rotated box. Points a and b represent the frictional force and torque required to balance the gravitational force and torque of the box.}
\label{LimitSurface}
\end{figure}

\begin{figure}
\centering
\includegraphics[scale=0.25]{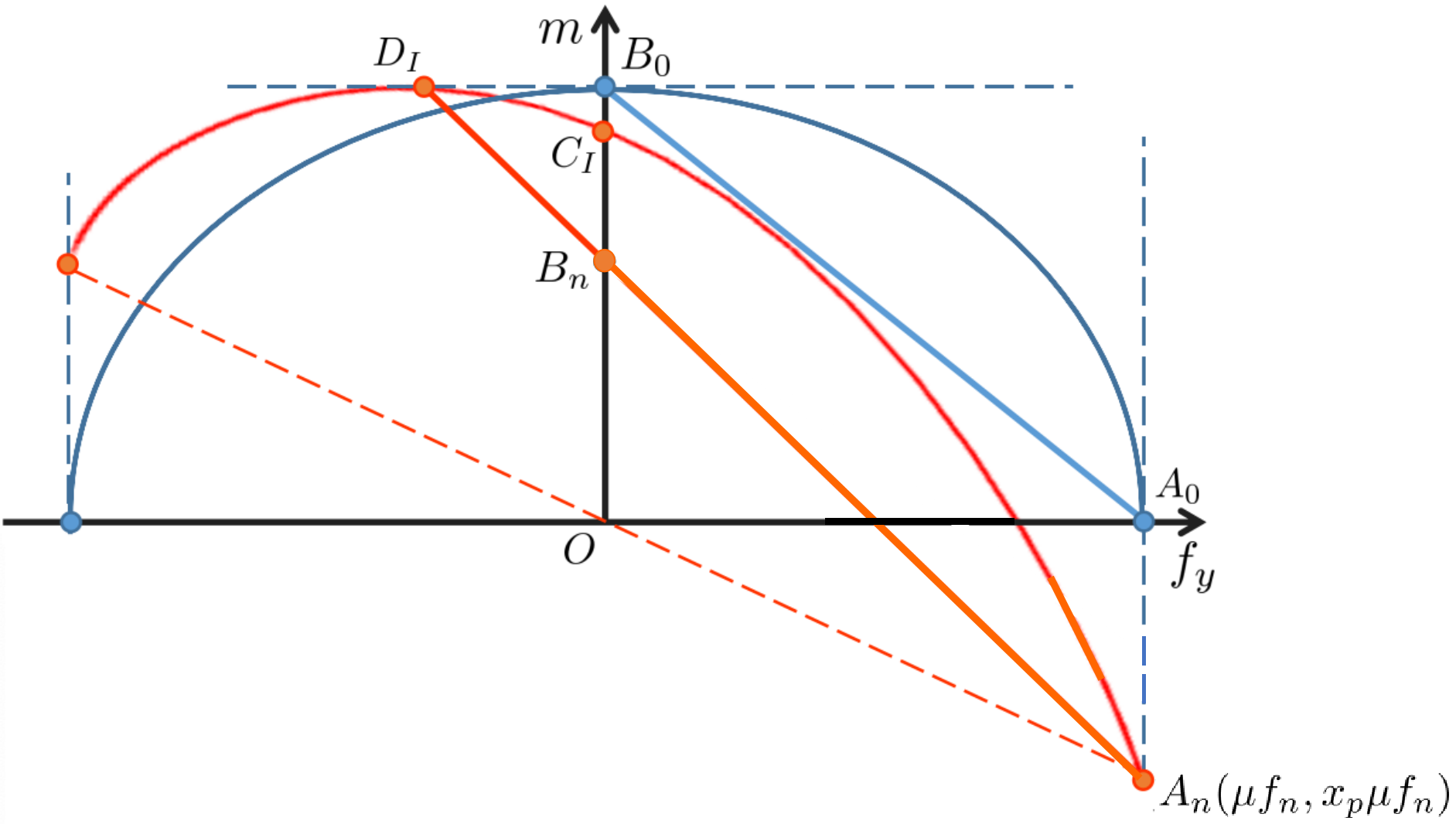}
\caption{The blue limit curve represents the case where the CoP is at the gripping pad's geometric center and the straight line connecting $A_{0}$ and $B_{0}$ is the control line. The red limit curve represents the case where CoP is slightly away from the gripping pad's geometric center where line $A_{n} D_{I}$ is constructed to obtain a new control line $A_{n}B_{n}$.}
\label{LC}
\end{figure}

When manipulating a box along a task space trajectory, the gripping force needs to be regulated based on the box's gravitational moment change. Since the friction force countering the box's gravity always points upwards, we can define a gripping coordinate with its origin at the gripper's geometric center and its y-axis pointing upwards and z-axis pointing towards the box. Therefore, the force along the $x$-axis always equals zero and the limit surface for the well-aligned case in Figure~\ref{ContactMechanics}d can be described using a limit curve (Figure~\ref{LimitSurface}b, left and Figure~\ref{LC}, blue curve) \cite{howe1996practical, lee1991fixture}.

According to Limit Surface Theory (Section~\ref{limit surface theory}), if a single pair of force and torque is applied to a slider, no slippage will occur if this pair is inside the limit curve. In our case, assuming the pair of gravitational force and moment for the initial box configuration (Figure~\ref{LimitSurface}a, left) is inside the limit curve (Figure~\ref{LimitSurface}b, left). When the robot rotates the box
 along a task space trajectory (Figure~\ref{LimitSurface}a, right), the gravitational moment increases due to the increased moment arm and the magnitude of gravity does not change. Therefore, the point in the limit curve moves upwards (Figure~\ref{LimitSurface}b, right). A new responsive limit curve needs to cover the updated point by increasing the gripping force (Figure~\ref{LimitSurface}b, right). Since the limit curve is always convex, we design a simplified control curve using a straight line that passes through the pair of gravitational force and momentum (Figure~\ref{LimitSurface}b, right). This control curve provides an extra safety boundary for slipping avoidance. Given the gravitational force and moment pair as $(f_{g},\tau_{g})$, the desired gripping force can be obtained by solving:
\begin{align}
    \frac{f_{g}}{\mu f_{n}} + \frac{\tau_{g}}{\tau_{\text{max}}(f_{n})} = 1
\end{align}
For our rigid pad with approximated uniformly distributed pressure, the maximum torque is given by:
\begin{align}\label{uniform torque}
\tau_{\text{max}} \approx 2/3\mu f_{n}R,
\end{align} where $R$ is the radius of the pad \cite{howe1996practical}. Therefore, the gripping normal force is given by:
\begin{align}
   f_{n} = \frac{f_{g}}{\mu} + \frac{3\tau_{g}}{2\mu R}.
\end{align}

Since our alignment control law aims to limit the distance between the CoP and the pad's geometric center within a certain bound (see Section~\ref{alignment control}), the CoP does not always perfectly locate at the gripping pad's geometric center. Due to the change in the pressure distribution, a new limit curve needs to be examined, which is shown as the red curve in Figure~\ref{LC} (see Section~\ref{contact mechanics} for contact with external moment). Accordingly, a new control line is required to replace the old control line ${A_0}{B_0}$. To determine the new control line, we first solve for $A_n$. When the box is sliding under pure translation, the friction force is $ f_{y}(A_{n})= \int\mu p(x,y)dA=\mu f_n $ and the moment is $m(A_{n}) =\int \mu xp(x,y) d A$, and the slope of the line $A_nO$ is given by:
\begin{align}
\frac{m(A_{n})}{f_{y}(A_{n})}= \frac{\int x p(x, y) d A}{\int p(x, y) d A} = x_{p}
\label{COP}
\end{align}
where $x_p$ is the measured CoP. Therefore,
\begin{align}
A_n=(\mu f_n, x_p{\mu}f_n).
\end{align}
Since the intersection point $C_{I}$ between the limit curve and the moment axis cannot be easily solved, we choose ${A_n}{B_n}$ as the new control line where $B_n$ is the intersection between line $A_nD_I$ and the moment axis, with $D_I$ the max moment point on the limit curve. Because the limit curve is convex \cite{drucker1954coulomb, goyal1989planar} and $B_n$ is under the $C_I$, the section $A_nB_n$ is also under the limit curve. As discussed in Section~\ref{limit surface theory}, $C_I$ does not change as long as the distance between CoP and the pad's geometric center is a constant. Therefore, to get an analytical solution for $B_n$ and $D_I$, we design an virtual CoP on the $x$-axis of the pad's local frame, $ {\bf P_{x}}=[|{\bf P}|, 0 ]^{T}$, where $\text{CoP}$, $\bf{P}$ is measured from force sensors. The averaged pressure $p_{0}$ in Equation \ref{pd} is given by $p_0 = f_n/(\pi R^2)$. Take $p_{0}$ into Equation \ref{COP} and rewrite the equation using polar coordinates parameterized by $(r, \theta)$:
\begin{align}
|{\bf P}|= \frac{1}{f_n}\int_{-\pi}^{\pi}\int_{0}^{R} r cos(\theta) (ar cos(\theta)+p_0) r \mathrm{d} r\mathrm{d} \theta
\end{align}
As a result,
\begin{align}
a = \frac{4f_n|{\bf P}|}{\pi R^4}
\end{align}
The max moment $D_I$ is obtained when CoR is located at the pad's geometric center \cite{howe1996practical}. Therefore, the Equation \ref{friction} written in terms of polar coordinates is given by: 
\begin{align}
f_{xD_{I}}=-\int_{-\pi}^{\pi}\int_{0}^{R} \mu r sin(\theta) (ar cos(\theta)+p_0)  \mathrm{d} r\mathrm{d} \theta = 0
\end{align}

\begin{align}
f_{yD_{I}}= \int_{-\pi}^{\pi}\int_{0}^{R} \mu r cos(\theta) (ar cos(\theta)+p_0)  \mathrm{d} r\mathrm{d} \theta = \frac{\pi R^3 a \mu}{3} 
\end{align}

\begin{align}
m_{D_{I}} =\int_{-\pi}^{\pi}\int_{0}^{R} \mu r^2 (ar cos(\theta)+p_0) \mathrm{d} r\mathrm{d} \theta = \frac{2\pi R^3\mu p_0}{3}
\end{align}
And $D_{I}$ can be solved as:

\begin{align}
D_I = \left (\frac{4 \mu f_n|{\bf P}|}{3R},  \frac{2 \mu f_n R}{3} \right)
\end{align}
As we can see, the moment element of $D_{I}$, which is the maximum moment of the limit curve share the same value as the limit curve in which no external moment is applied (Equation \ref{uniform torque} and Figure~\ref{LC}). 
Finally, we have
\begin{align}
B_n = \left(0, 2\mu f_n\left(\frac{2|{\bf P}|^2-R^2}{4|{\bf P}|-3R}\right)\right).
\end{align}

In our gripping force controller, a new control line is generated in each control loop according to the measured CoP position.

\subsection{Task Space Trajectory Tracking}
\begin{figure}[ht]
\centering
\includegraphics[width=0.70\linewidth]{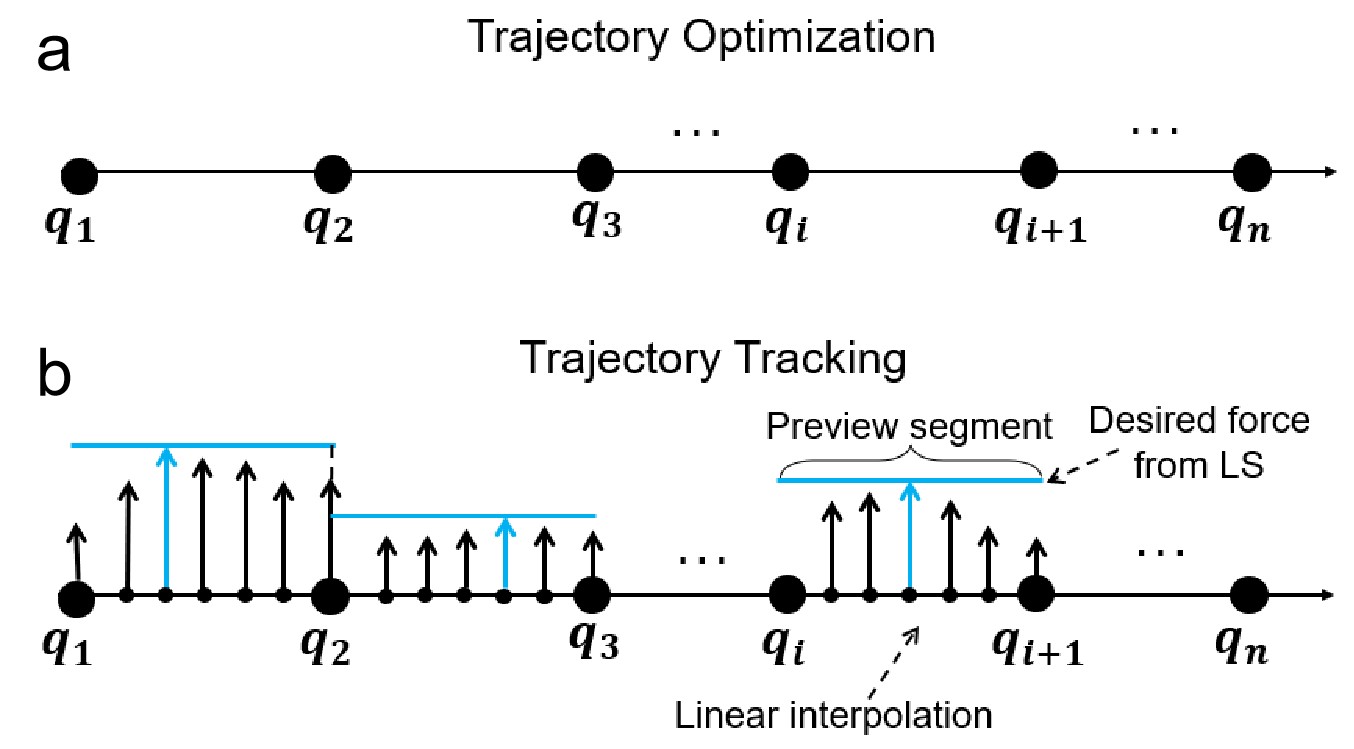}
\caption{(a) Generating trajectories in robot configuration space by trajectory optimization. (b) The reference force within two adjacent nodes along the trajectory is acquired by first applying linear interpolation between adjacent nodes and then selecting the local maximum force estimated using limit curves.}
\label{LimitSurface2}
\end{figure}

After generating a task space trajectory, a detailed gripping force controller is also required for trajectory tracking. Supposing the robot is at the $i$th node of the trajectory (Figure~\ref{LimitSurface2}a), a linear interpolation is first implemented between the $i$th and the $(i+1)$th node to improve control resolution (Figure~\ref{LimitSurface2}b). Then the robot observes all the discretized configurations within the $i$th and the $(i+1)$th nodes to find the maximum required normal force, which is estimated from the limit curve based on the box's CoM position (Figure~\ref{LimitSurface2}b). Then the acquired upper bound normal force becomes the reference of all the discretized control nodes between the $i$th and the $(i+1)$th node. Using upper-bound reference is a safe strategy for our low-cost platform with low control frequency. Since the traditional Jacobian-based inverse kinematics method can not easily include different constraints, such as joint limit and collision \cite{fallon2015architecture}, a nonlinear programming method (NLP) is used to solve for a desired joint configuration by adjusting the pad's position and orientation based on the sensory feedback and the limit curve force estimation. The NLP is implemented for each control node between two adjacent joint configuration nodes along the trajectory, which is formulated as:
\begin{align}
\underset{\tilde{{\bf  q}[k]}}{\text{minimize}} \quad & J = ||{\bf  \tilde{q}}[k]-{\bf  q}[k]||^{2} & \text{(Cost function)}\\
\textrm{subject to:} \quad 
&{\bf  T_{L}}(\tilde{{{\bf  q}[k]}}) - {\bf  T_{L}}({\bf  q}[k]) = [0,+\Delta s[k-1],0]^{\mathsf{T}} &  \text{(L force)}\\
&{\bf  T_{R}}(\tilde{{\bf  q}[k]}) - {\bf  T_{R}}({\bf  q}[k]) = [0,-\Delta s[k-1],0]^{\mathsf{T}} &  \text{(R force)}\\
&{\bf  R_{L}}(\tilde{{\bf  q}[k]}) = {\bf  R_{L}}({\bf  q}[k])e^{{\bf  \hat{n}_{L}}[k-1]\Delta \theta_{L}[k-1]} &  \text{(L CoP)}\\
&{\bf  R_{R}}(\tilde{{\bf  q}[k]}) = {\bf  R_{R}}({\bf  q}[k])e^{{\bf  \hat{n}_{R}}[k-1]\Delta \theta_{R}[k-1]} &  \text{(R CoP)}\\
&\Delta s[k-1] = K_{s}\left(f_{n}[k-1] - \left(\frac{f_{g}[k]}{\mu} + \frac{3\tau_{g}[k]}{2\mu R}\right)\right) &  \text{(LS Estimation)} \label{LS update}\\ 
&\text{kinematics constraints} 
\end{align}
where $\tilde{{\bf  q}[k]}$ and ${\bf  q}[k]$ are the desired joint configuration and the discretized configuration for each control node. Index $[k-1]$ indicates the measurements of pad's distance or orientation from the previous control node. The gripping force is updated based on limit surface estimation (Equation \ref{LS update}). The algorithm for the task space trajectory tracking is shown in Algorithm~\ref{traj_alg}:
\begin{algorithm}[ht]
\caption{Gripping control in task space} \label{traj_alg}
\begin{algorithmic}
\State Initialization:
\State ${\bf q[1]} \gets {\bf q}_{\text{initial}}$
\State ${\bf q[N]} \gets {\bf q}_{\text{goal}}$
\State Run trajectory optimization: ${\bf Q} \gets [{\bf q}[1],{\bf q}[2],\hdots, {\bf q}[N]]$
\State Start trajectory tracking:
\State $k \gets 0$
\While{$k < N$}
    \State $k \gets k+1$
    \State ${\bf Q_{L}} = \text{linear interpolate} \left \{ {\bf q}[k], {\bf q[k+1]} \right \} \in {\bf Q}$
    \State $\tau_{\text{max}} \gets \text{search} {\bf Q_{L}}$
    \State $f_{\text{ref}} \gets \text{based on limit curve}$
    \State $s \gets 0$
    \While{$s < m$}
        \State $s \gets s+1$
        \State ${\bf \tilde{q}[k][s]} \gets \text{Run hybrid controller based on} {\bf {q}[k][s]}$
        \State ${\bf q_{\text{robot}}} \gets {\bf \tilde{q}[k][s]}$ \text{(Robot moving)}
    \EndWhile
\EndWhile
\end{algorithmic}
\end{algorithm}

\section{Experiments} \label{exp}
\subsection{Experimental Design}
Four experiments are implemented to test our force-sensing grippers and the hybrid position-alignment-force controller. The experimental details are introduced as follows:

\noindent \textbf{Experiment 1:} 
The robot is commanded to move a box with a 100g mass along the horizontal direction from the robot's middle to its right and then moves back to its middle (Figure~\ref{TaskDemo}a, black arrows). The robot tracks a 2N gripping force with our position-alignment-force controller. The task space trajectory is preplanned. \\
\noindent \textbf{Experiment 2:} 
The robot is commanded to move a box with a 100g mass along the vertical direction while changing the box's orientation (Figure~\ref{TaskDemo2}a, black arrows). The box's CoM location relative to its local frame is a priori known and the box's position relative to the world frame is estimated by tracking the tag attached to the top of the box. The initial gripping reference is 2N. The robot regulates the gripping force according to the limit surface estimation. \\
\noindent \textbf{Experiment 3 and 4}: In these two experiments, the robot is commanded to first grip and then lift a box with vertical sides (Experiment 3, Figure~\ref{Vertical}, top) and a box with 15$^\circ$ tilted sides (Experiment 4, Figure~\ref{Tilted}, top). A simple force integral controller which only tracks the desired reference gripping normal force and our hybrid force alignment controller are compared. The initial gripping orientation is designed to have a $10^\circ$ misalignment with the side faces of the boxes to test the significance of surface alignment in the control process. During the gripping phase, the grippers gradually approach the sides of the box and grip them till reach desired force or force and CoP references for the two controllers to be compared. Then the robot lifts the box with the controllers keep running till the box is fully lifted above the ground. 

\subsection{Experimental Results} \label{Exp Result}
\begin{figure}[H]
\centering
\includegraphics[width=1.0\linewidth]{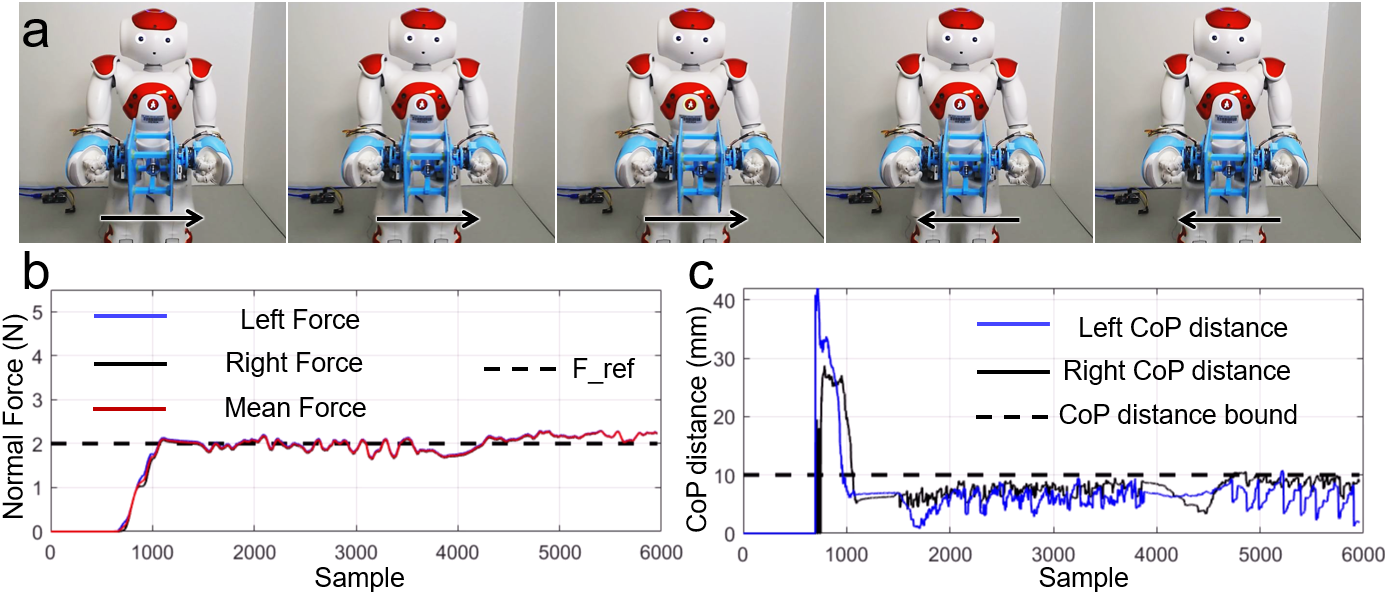}
\caption{(a) Snapshots of the NAO moves a box with 100g mass along the horizontal direction. (b) and (c) show the normal force and CoP distance in the movement.}
\label{TaskDemo}
\end{figure}

\begin{figure}[H]
\centering
\includegraphics[width=1.0\linewidth]{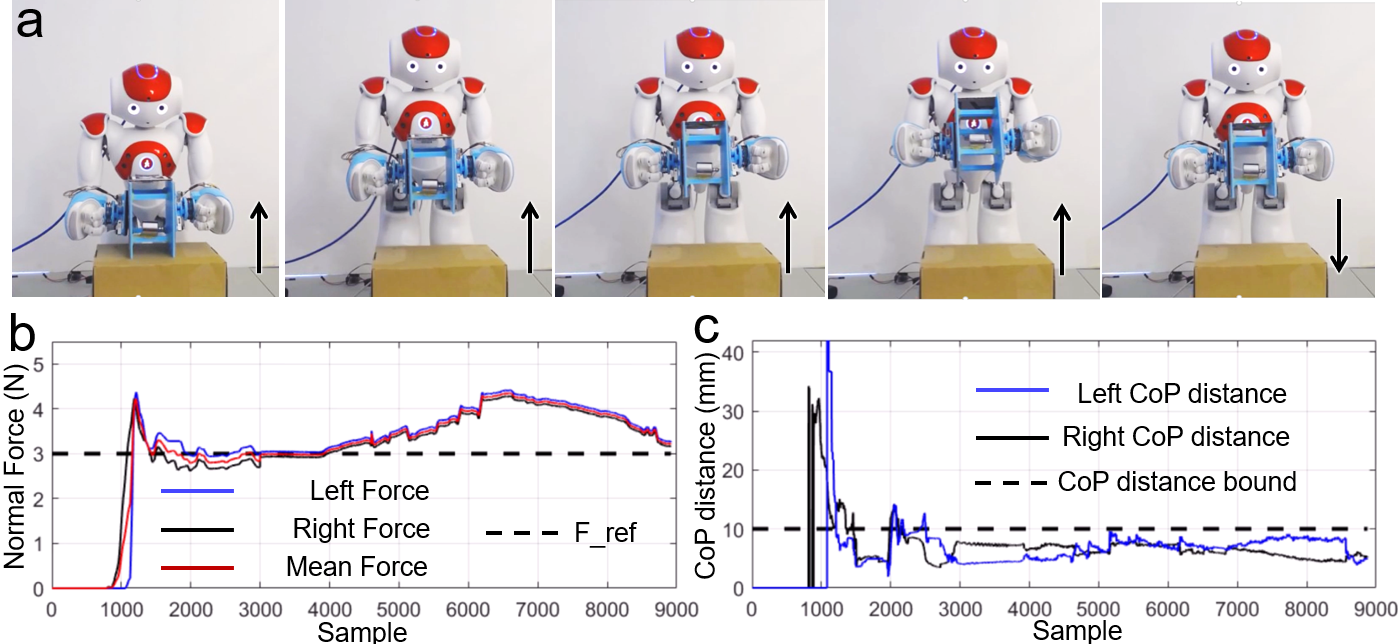}
\caption{(a) Snapshots of the NAO moves a box with 100g mass along the vertical direction. (b) and (c) show the normal force and CoP distance in the movement.}
\label{TaskDemo2}
\end{figure}

\begin{figure}[H]
\centering
\includegraphics[width=1.0\linewidth]{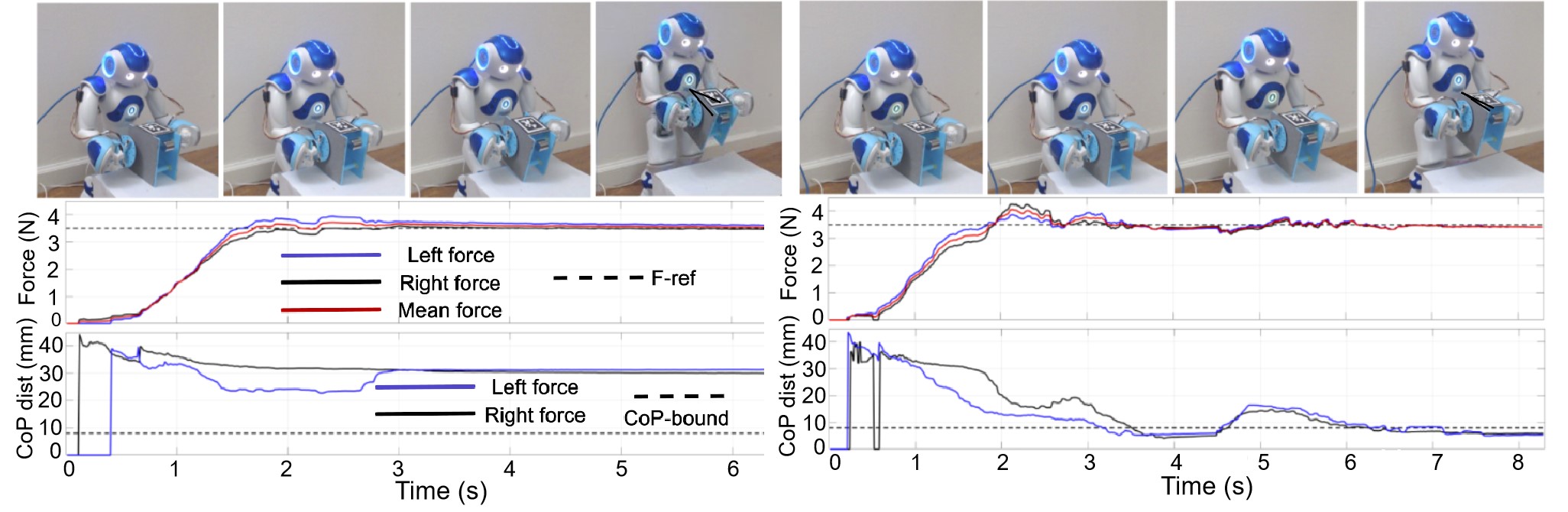}
\caption{Comparison between simple gripping force control and hybrid force alignment control for lifting a box. Top shows snapshots of the NAO lifts a box with 200g from ground. Left represents gripping force control; right represents hybrid gripping force alignment control. Bottom shows gripping force and CoP of the left (blue) and right (black) grippers and their mean gripping force (red). The dashed lines show the reference gripping force and the CoP bound.}
\label{Vertical}
\end{figure}

\begin{figure}[H]
\centering
\includegraphics[width=1.0\linewidth]{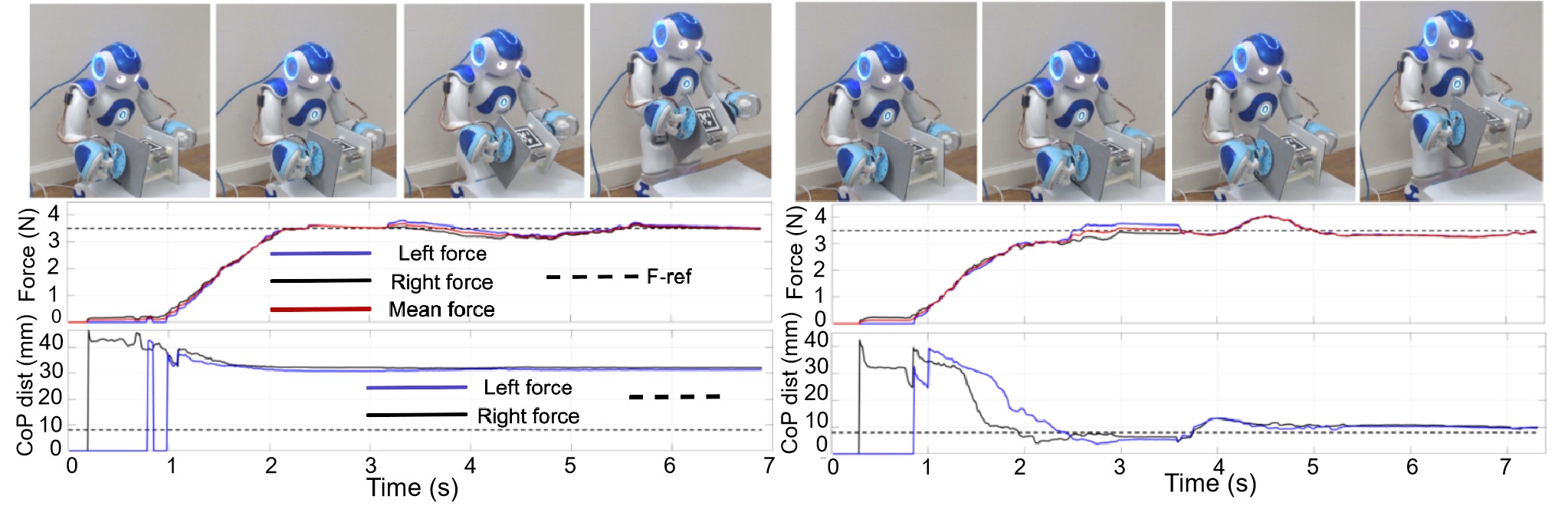}
\caption{Comparison between simple gripping force control and hybrid force alignment control for lifting a box with 15$^\circ$ tilted side faces. Top shows snapshots of the NAO lifts a box with 200g from ground. Left represents gripping force control; right represents hybrid gripping force alignment control. Bottom shows gripping force and CoP of the left (blue) and right (black) grippers and their mean gripping force (red). The dashed lines show the reference gripping force and the CoP bound.}
\label{Tilted}
\end{figure}

\textbf{Experiment 1}: The result shows the force controller tracks the desired force stably along the trajectory (Figure~\ref{TaskDemo}b). The CoP distance (Figure~\ref{TaskDemo}c), defined as the distance between the measured CoP and the gripper's geometric center, is controlled within a designed threshold of 10 mm (Figure~\ref{TaskDemo}c).\newline
\textbf{Experiment 2}: The result shows that the measured force profile forms a gradually increased stair-like shape moving upwards when the gravitational moment increases and decreases as the gravitational moment decreases (Figure~\ref{TaskDemo2}b). Additionally,  the CoP distance is well limited within the 10 mm designed threshold along the trajectory (Figure~\ref{TaskDemo2}c). \newline
\textbf{Experiment 3 and 4}: In both experiments, the boxes gripped by the simple normal force integral controller slip significantly more than the box gripped by our hybrid force-alignment-controller (Figure ~\ref{Vertical} and \ref{Tilted}, top, right). Especially for the challenging case where the box has tilted sides, the box slips around a point inside the gripper (Figure ~\ref{Tilted}, top). It is clearly shown that the normal forces are tracked well for both controllers with the major difference being the CoP distance (Figure ~\ref{Vertical} and \ref{Tilted}, bottom). Our hybrid controller takes into account the CoP distance which ensures better contact between the gripping pads and the surfaces of the box resulting in slippage avoidance.

The overall experiments demonstrate that our proposed control method enables stable dual-arm manipulation for box-like objects in task space for smaller-sized and low-cost dual-arm robots.

\section{Conclusions} \label{sec_3.6}
This {paper} introduces the design, calibration, and control method of a pair of force-sensing grippers for manipulating box-like objects for smaller-sized or low-cost dual-arm platforms. The grippers measure normal force and contact CoP accurately, benefiting from the calibration process. The compliant design of the grippers enhances the gripping stability. Unlike the traditional position-force controller that highly depends on the manipulator's accurate kinematic control, {the designed close-loop alignment controller incorporating CoP measurement significantly increases gripping stability for manipulators with low kinematic accuracy.} The limit surface theory is applied to our control framework to predict the gripping forces when moving the object along a task-space trajectory, which is novel in the literature on dual-arm manipulations. 

The presented force-sensing grippers are developed for gripping flat surfaces. Due to the limited contact area, the current design may be challenging to grip curved surfaces. A deformable or flexible gripper design, together with our compliant spring mechanism, may be effective for the stable manipulation of objects with different shapes. The effectiveness of our proposed gripping control method has been demonstrated by moving the objects in different tasks. Our future work will include more complex manipulation tasks, including three-DoF rotation, which requires a more comprehensive gripping force estimation based on contact modeling.

\section{Acknowledgement}
This work was supported by NUS Startup grants A-0009059-02-00 and A-0009059-03-00, CDE Board account E-465-00-0009-01, SMI Grant A-8000081-02-00, MOE Tier 2 grant A-8000424-00-00, and National Research Foundation, Singapore, under its Medium Sized Centre Programme - Centre for Advanced Robotics Technology Innovation (CARTIN), sub award A-0009428-08-00.
 
\bibliographystyle{IEEEtran}
\bibliography{JMR.bib}
\end{document}